\documentclass[runningheads]{llncs}
\usepackage[T1]{fontenc}
\usepackage{subcaption}  

\usepackage{graphicx,verbatim, xcolor}
\usepackage{amsmath}
\usepackage{hyperref}
\usepackage{booktabs} 
\usepackage{multirow}
\begin{document}

\title{CXR-TFT: Multi-Modal Temporal Fusion Transformer for Predicting Chest X-ray Trajectories}
\titlerunning{CXR-TFT}
%
\author{Mehak Arora\inst{1}\thanks{Corresponding Author: mehak.arora@duke.edu}
\and Ayman Ali\inst{1}
\and Kaiyuan Wu\inst{1}
\and Carolyn Davis\inst{2}
\and Takashi Shimazui\inst{2}
\and Mahmoud Alwakeel\inst{1}
\and Victor Moas\inst{1}
\and Philip Yang\inst{2}
\and Annette Esper\inst{2}
\and Rishikesan Kamaleswaran\inst{1}
}
\authorrunning{M. Arora et al.}
%
\institute{Duke University, Durham NC 27708 \email{\{mehak.arora, ayman.ali, vincent.wu, mahmoud.alwakeel, victor.moas, r.kamaleswaran\}@duke.edu} \\ \and 
Emory University, Atlanta GA 30322 \\ 
\email{cmydavis@gmail.com,\{tshima2,philip.yang,aesper\}@emory.edu}}

%

\maketitle              
\begin{abstract}
In intensive care units (ICUs), patients with complex clinical conditions require vigilant monitoring and prompt interventions. Chest X-rays (CXRs) are a vital diagnostic tool, providing insights into clinical trajectories, but their irregular acquisition limits their utility. Existing tools for CXR interpretation are constrained by cross-sectional analysis, failing to capture temporal dynamics. To address this, we introduce CXR-TFT, a novel multi-modal framework that integrates temporally sparse CXR imaging and radiology reports with high-frequency clinical data—such as vital signs, laboratory values, and respiratory flow sheets—to predict the trajectory of CXR findings in critically ill patients. CXR-TFT leverages latent embeddings from a vision encoder that are temporally aligned with hourly clinical data through interpolation. A transformer model is then trained to predict CXR embeddings at each hour, conditioned on previous embeddings and clinical measurements. In a retrospective study of 20,000 ICU patients, CXR-TFT demonstrated high accuracy in forecasting abnormal CXR findings up to 12 hours before they became radiographically evident. This predictive capability in clinical data holds significant potential for enhancing the management of time-sensitive conditions like acute respiratory distress syndrome, where early intervention is crucial and diagnoses are often delayed. By providing distinctive temporal resolution in prognostic CXR analysis, CXR-TFT offers actionable 'whole patient' insights that can directly improve clinical outcomes.


\end{abstract}
\section{Introduction}

Patients that require intensive care unit (ICU) level of care generally have complex and diverse clinical pathologies that require careful monitoring and timely intervention. Portable chest radiographs (CXRs) are the most requested imaging in ICU patients for a variety of reasons: they are rapid to obtain, can be done bedside (critical for unstable patients), are used to evaluate support devices and lines, and can provide important diagnostic information, particularly for pulmonary pathology. \cite{toy2022imaging} Importantly, there are various conditions that are first recognized or diagnosed with CXRs, like a consolidation indicative of a pneumonia, new pleural effusions in the setting of volume overload, or pulmonary edema. \cite{laroia2021acr} These conditions often develop as complications related to the ICU patients' underlying pathology, and can each carry significant morbidity and mortality, like acute respiratory distress syndrome (ARDS). \cite{bellani2020missed} For most of these pathologies, early recognition and intervention is critical to improving outcomes. \cite{konig2017precision} 

However, many contemporary machine learning models that are applied in the ICU setting---like cohort phenotyping or outcome prediction---either only leverage radiology reads of CXRs and/or do not use imaging data all-together \cite{gutierrez2020artificial,van2021moving} 
This has a few notable limitations, primarily that CXRs contain valuable information that influences clinical decision making and subsequent patient trajectories, and that the radiology report of the CXR alone may be delayed and may not convey information that is either implied or acted on prior the time of the read. Therefore, there is an important need to better integrate imaging into clinical machine learning projects, particularly those in the ICU as many outcomes are associated with disease trajectories partially reflected in CXR data. 
Independently, there has been significant research on using machine learning for CXR interpretation \cite{ahmad2023machine} as well as CXR generation\cite{bluethgen2024vision}. 
Foundational medical imaging models \cite{stevens2024bioclip} \cite{cxr-bert} \cite{xu2023elixr} have been successful in learning rich representations of CXR image data, enabling data-efficient training for downstream tasks like abnormality classification. Also, models that incorporate longitudinal CXR data have been shown to outperform models that are restricted to cross-sectional CXR analysis. \cite{gu2023biomedjourneycounterfactualbiomedicalimage},\cite{bannur2023learning}, \cite{kyung2024towards}. 

In this study, we applied recent advancements in CXR image analysis to a cohort of ICU patients, hypothesizing that the most likely CXR could be estimated at any point during a patient’s ICU stay. This capability is particularly significant for decompensation models, potentially shortening the time to clinical intervention. To accomplish this, we developed CXR-TFT (Chest X-ray Temporal Fusion Transformer), a transformer-based model that integrates hourly clinical measurements—such as lab values, vital signs, and ventilator parameters—with previous CXR embeddings to predict the most probable CXR representation in latent space. The latent embedding space of a vision-language model is continuous\cite{mikolov2013efficient} and imbued with semantic meaning\cite{stevens2024bioclip}. This allows for interpolation between embeddings, helping us overcome the challenge of temporally aligning information from multi-modal irregularly sampled time series and forming the key technical contribution of this work. We hypothesize that this 'whole person' approach to characterizing acute clinical physiology, allowing for a more robust characterization of the multi-modal latent representation, enabling richer and deep fidelity in the generated images. 

\begin{figure}[t]
    \centering
    \includegraphics[trim=0 3.5cm 0 3cm, clip,width=\linewidth]{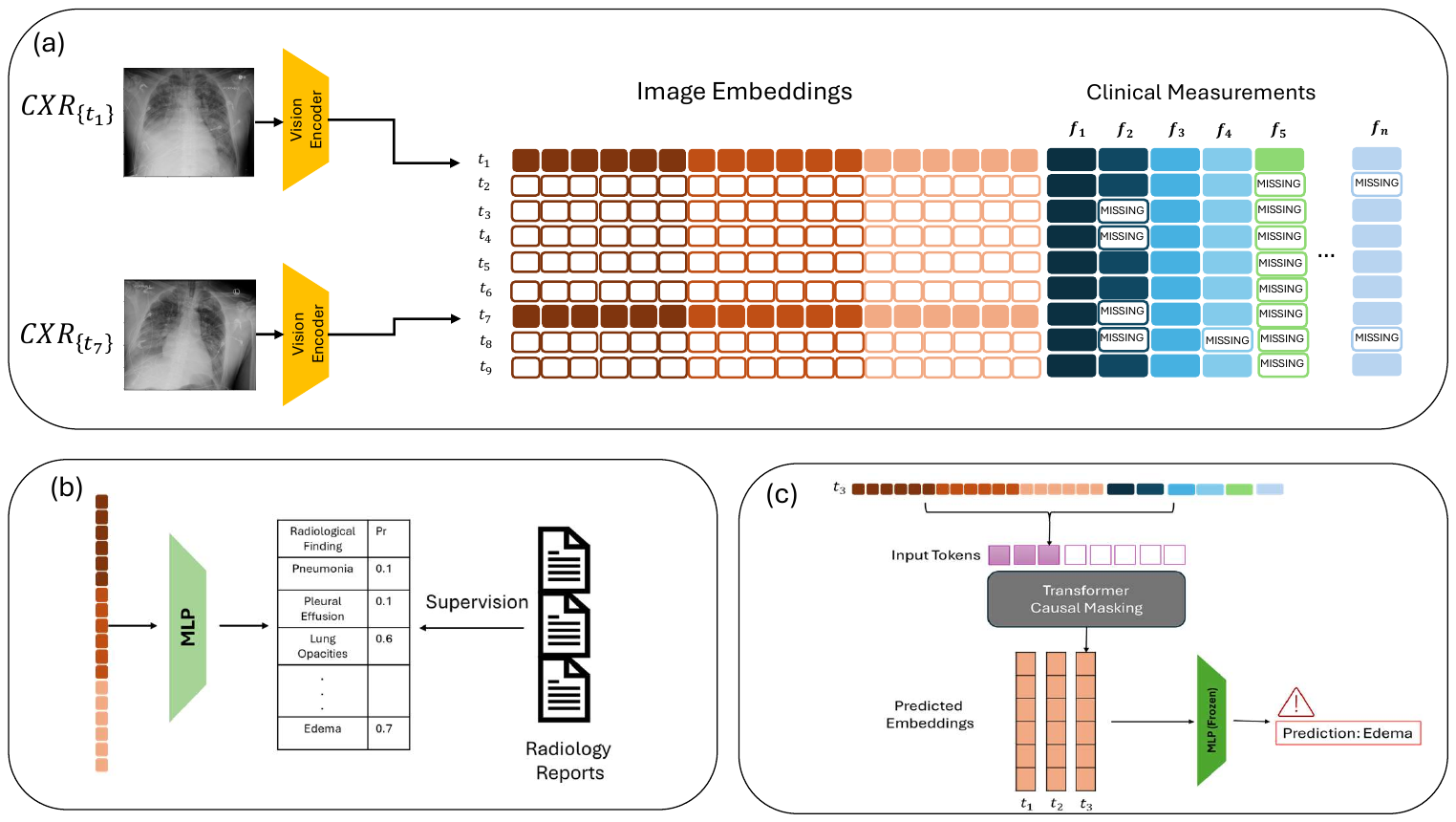}
    \caption{Visual Depiction of the CXR-TLT Framework.(a) Sparsely recorded CXR images and irregularly sampled clinical measurements are concatenated at the input to the transformer model, (b) A Multi-layer Perceptron is trained to detect radiographic findings from embeddings of the vision encoder, with ground truth supervision from radiology reports, (c) CXR-TLT estimates future CXR embeddings which can predict the likelihood of radiographic findings before they are seen on subsequent CXRs.}
    \label{fig1}
\end{figure}

\section{Methods}
A high-level overview of our trajectory estimation framework is shown in Figure \ref{fig1}
\subsection{The Proposed Framework}
At any time $t_k$ during a patient's stay in the ICU, given a sequence of clinical measurements $F = \{F_{t_0}, F_{t_1},F_{t_2} ..., F_{t_{k-1}}\}$, where $F_t = [f^1_t, f^2_t, ..., f^n_t ]$ is a vector of $n$ clinical features under consideration, and given a sequence of sparsely sampled, previously recorded CXR images $I^{p} = \{I^p_{t_0}, (\bullet) ,I^p_{t_2} ..., I^p_{t_{k-1}}\}$ where $I^p_{t} = encoder_{vision}(CXR_{t})$ is the latent embedding representation of a CXR image obtained via a pretrained vision encoder, and $(\bullet)$ represents time points with no recorded CXR scans, the proposed model learns to predict $I^E_{t_k}$, the estimated CXR embedding at time $t_k$. The target output sequence $I^T = \{I^T_{t_0},I^T_{t_0}  ,I^T_{t_2} ..., I^T_{t_{k-1}}\}$ used to train the model is obtained by linear interpolation in the embedding space between two recorded CXRs. Concretely, if a CXR scan was performed at time $t_{k_1}$, and the next CXR scan was performed at $t_{k_2}$, then

\begin{equation}
\label{interp}
I^T_{t_{k'}} = 
\begin{cases}
I^T_{k_1} = \text{encoder}_{vision}(CXR_{t_{k_1}}), & \text{if } k' = k_1 \\
I^T_{k_2} = \text{encoder}_{vision}(CXR_{t_{k_2}}), & \text{if } k' = k_2 \\
\frac{I^T_{k_2} - I^T_{k_1}}{k_2 - k_1} \times (k' - k_1) + I^T_{k_1}, & \text{if } k_1 < k' < k_2
\end{cases}
\end{equation}

\subsection{Dataset Preparation}
This is a single-center retrospective cohort study at an academic institution. Included were all adult patients admitted to any ICU between January 2015 to December 2021 who had more than one CXR performed during their hospitalization. A total of 17,690 patients met criteria, for which we extracted all single-view anteroposterior (AP) frontal chest radiographic images and their corresponding radiology reports. We also extracted demographic information and clinical measurements like vitals, laboratory values, ventilator flowsheet information, and aggregated them into hourly intervals across the entire ICU length of stay.

\subsection{Data Preprocessing}

\subsubsection{Clinical Measurements} \label{sec:ehr}
All clinical measurements from the Electronic Medical Record (EMR) are were organized into hourly bins. For variables with multiple recordings within an hour, values were first validated against physiologically possible bounds (determined through clinician consultation), with out-of-range values discarded and the remaining values averaged. Numerical values were min-max normalization using healthy patient reference ranges. Missing clinical measurements were handled with forward-fill imputation. If no recorded value existed, missing values were imputed using the median of the normal (healthy) range of values. Categorical variables (gender, ICU type, etc.) were one-hot encoded. This processing resulted in a clinical feature vector $F_t = [f^1_t, f^2_t, ..., f^n_t ]$ with $n = 82$.


\subsubsection{Image Encoding} \label{sec:image-encoder}
BioCLIP \cite{stevens2024bioclip}, a vision language model trained to align radiology reports with corresponding image embeddings, was used to extract the latent space representation $I_{t_k} \in \mathcal{R}^512$ of a chest x-ray image at time $t_k$. Target output sequences were generated by linear interpolation of successive CXR embeddings, as described in equation \ref{interp}. Data preceding the first recorded CXR and following the last recorded CXR was excluded for training and evaluation. To facilitate training, missing values, $(\bullet)$ in the previous CXR sequence $I^{p} = \{I^p_{t_0}, (\bullet) ,I^p_{t_2} ..., I^p_{t_{k-1}}\}$ were handled using forward-fill imputation.

\subsubsection{Radiology Reports} \label{sec:reports}
Radiology reports were used to provide a supervision signal to train a downstream classifier to predict radiological findings from image embeddings. We derived 10 classes of radiological findings from the text reports: cardiovascular findings ('Cardiomegaly'); pulmonary abnormalities ('Lung Opacity', 'Edema', 'Consolidation', 'Pneumonia', 'Atelectasis', 'Pneumothorax'); pleural abnormalities ('Pleural Effusion', 'Pleural Other'), and 'No Finding'. This was done using the CheXPert labeler tool \cite{irvin2019chexpert}.


\subsubsection{Modality Fusion} \label{sec:modality-fusion}
The input data to the transformer model at time $t_k$ is $X_{t_k} = [F_{t_k}^T, {I^p_{t_k}}^T]$, was a $594 \times 1$ vector formed by the concatenation of current clinical features and the latent embedding of the previously recorded CXR.

\subsection{Training CXR-TFT} \label{sec:training}
We trained an encoder-decoder transformer model\cite{vaswani2017attention} with a pre-norm architecture, an initial learning rate of $5e-4$, and the AdamW optimizer with a weight decay of $0.01$. Gradient clipping were used to prevent exploding gradients. The model was trained for $100$ epochs with an early stopping patience of $10$ epochs based on validation loss. We used a batch size of $32$ and a cosine learning rate scheduler with warmup for the first 10\% of training steps. To prevent overfitting, we applied dropout with a rate of $0.1$ throughout the network. The mean squared error (MSE) loss between the target CXR embeddings and the decoder outputs was used as the primary optimization objective. The code to the complete data processing and training setup can be found at our \hyperlink{https://github.com/Kamaleswaran-Lab/cxrgen}{Github Repository}.

\subsection{Classifier Regularization} \label{sec:mimic-classifier}
To improve the learning process, we trained a lightweight multilayer perceptron to predict key radiological findings using labels derived from radiology reports (Section \ref{sec:reports}). This model used embeddings from the BioCLIP vision encoder as input and labels from radiology reports at the output. It was trained on the MIMIC-CXR dataset \cite{johnson2019mimic}, which consists of over 200,000 CXR images with associated radiology reports.

The cross-entropy loss between predicted labels of the decoder output and the target labels was added to the training objective. This regularization encourages the predicted trajectories to align with our primary objective: accurately forecasting the likelihood of abnormal findings on CXR images. For $N$ training samples and $C$ classes radiological findings, each with sequence length $T_i$ where $i \in \{0, 1, \dots, N \}$, the training objective is given by Equation \ref{eq:obj}, where $y_{i,c,t}  = \text{MLP}(I^T_{t})$ and $p_{i,c,t} = \text{MLP}(I^p_{t})$ and $\theta$ are the parameters of our model. For our model, $C = 10$ and $\alpha = 0.5$.  

\begin{equation}
    \mathcal{L}_{MSE}(\theta) = \frac{1}{N} \sum_{i= 1}^{N} \frac{1}{T_i} \sum_{t= 1}^{T_i} (1 - \alpha) \left\lVert I^p_{t} - I^T_{t} \right\rVert_{2}^2
\end{equation}

\begin{equation}
    \mathcal{L}_{\text{BCE}}(\theta) = -\frac{1}{N} \sum_{i=1}^{N} \sum_{t= 1}^{T_i} \sum_{c=1}^{C} \left[ y_{i,c,t} \log(p_{i,c,t}) + (1-y_{i,c,t}) \log(1-p_{i,c,t}) \right]
\end{equation}

\begin{equation} \label{eq:obj}
     \mathcal{L}(\theta) = (1 - \alpha)L_{MSE}(\theta) + \alpha L_{BCE}(\theta) 
\end{equation}

\section{Results on Predicting Radiographic Findings from Predicted Embeddings}
\begin{figure}[t]
    \centering
    \begin{subfigure}{\textwidth}
        \centering
        \includegraphics[width=0.8\textwidth]{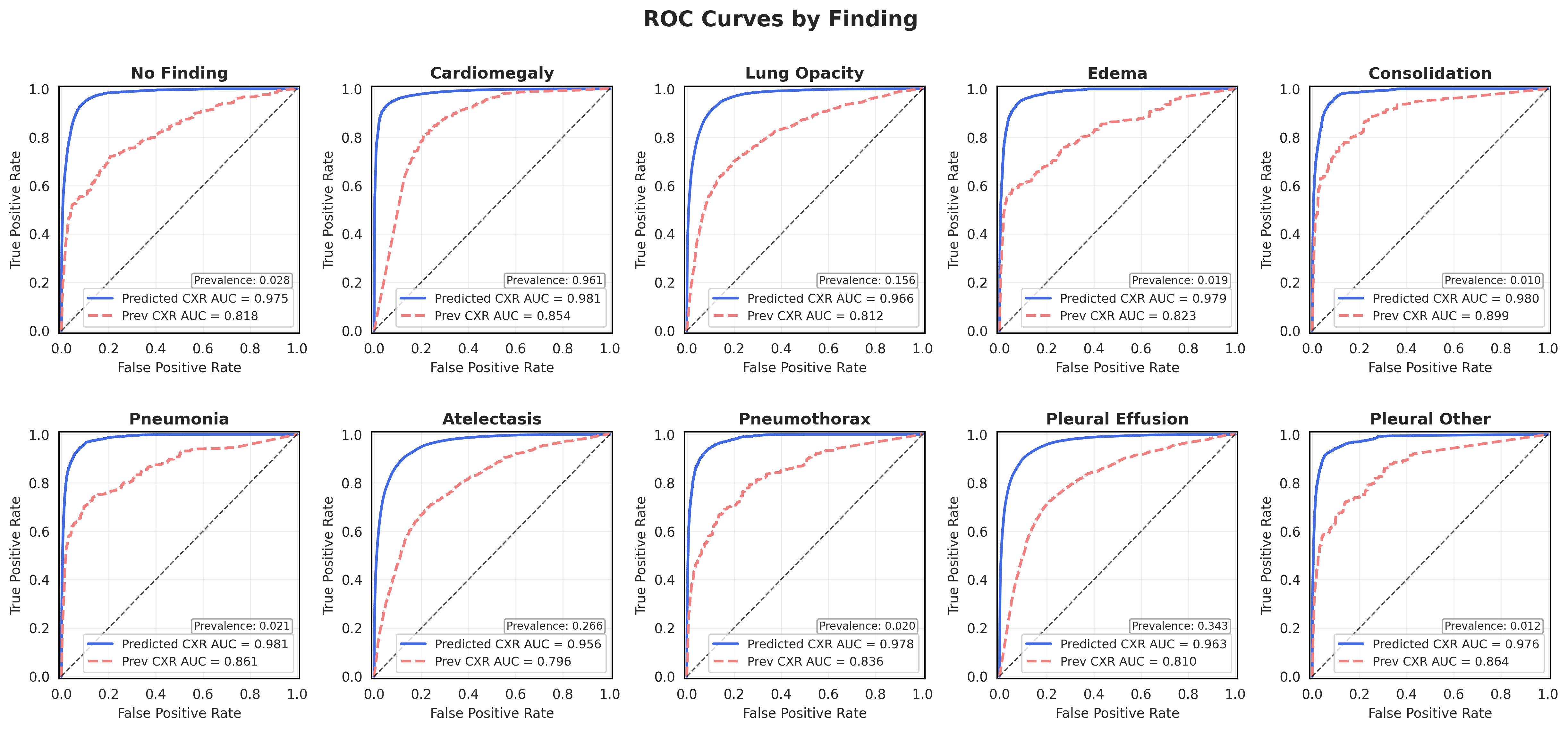}
        \caption{Receiver Operating Characteristic Curves}
    \end{subfigure}    
    \begin{subfigure}{\textwidth}
        \centering
        \includegraphics[width=0.8\textwidth]{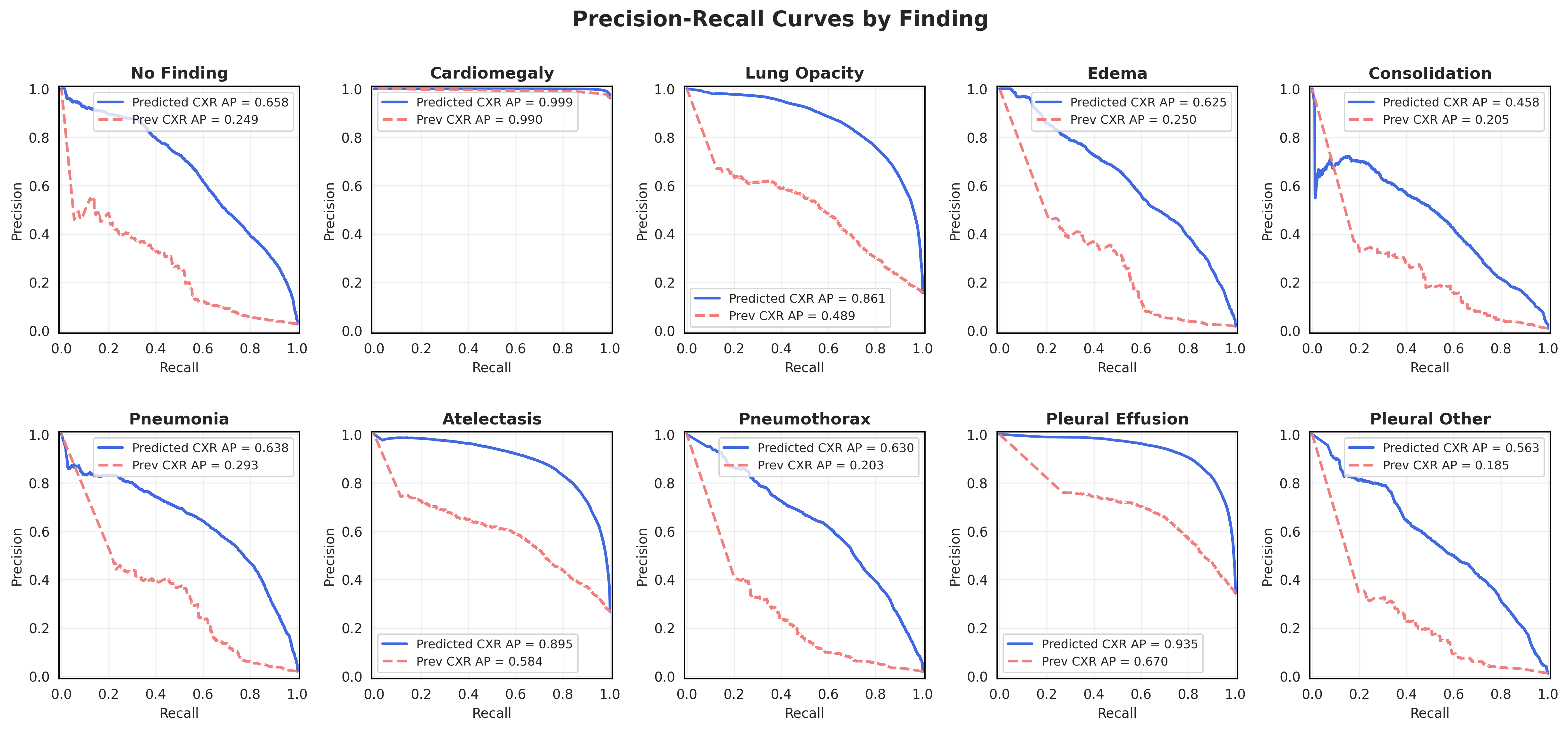}
        \caption{Precision Recall Curves}
    \end{subfigure}
    
    \caption{Performance comparison of detecting radiographic findings on the embeddings predicted by the transformer model, and the baseline of the previously recorded CXR. Figure (a) also denotes the prevalence of each class in the test set. }
    \label{fig:roc_prc}
\end{figure}

\begin{figure}[h]
    \centering
    
    \begin{subfigure}{0.48\textwidth}
        \centering
        \includegraphics[width=\textwidth]{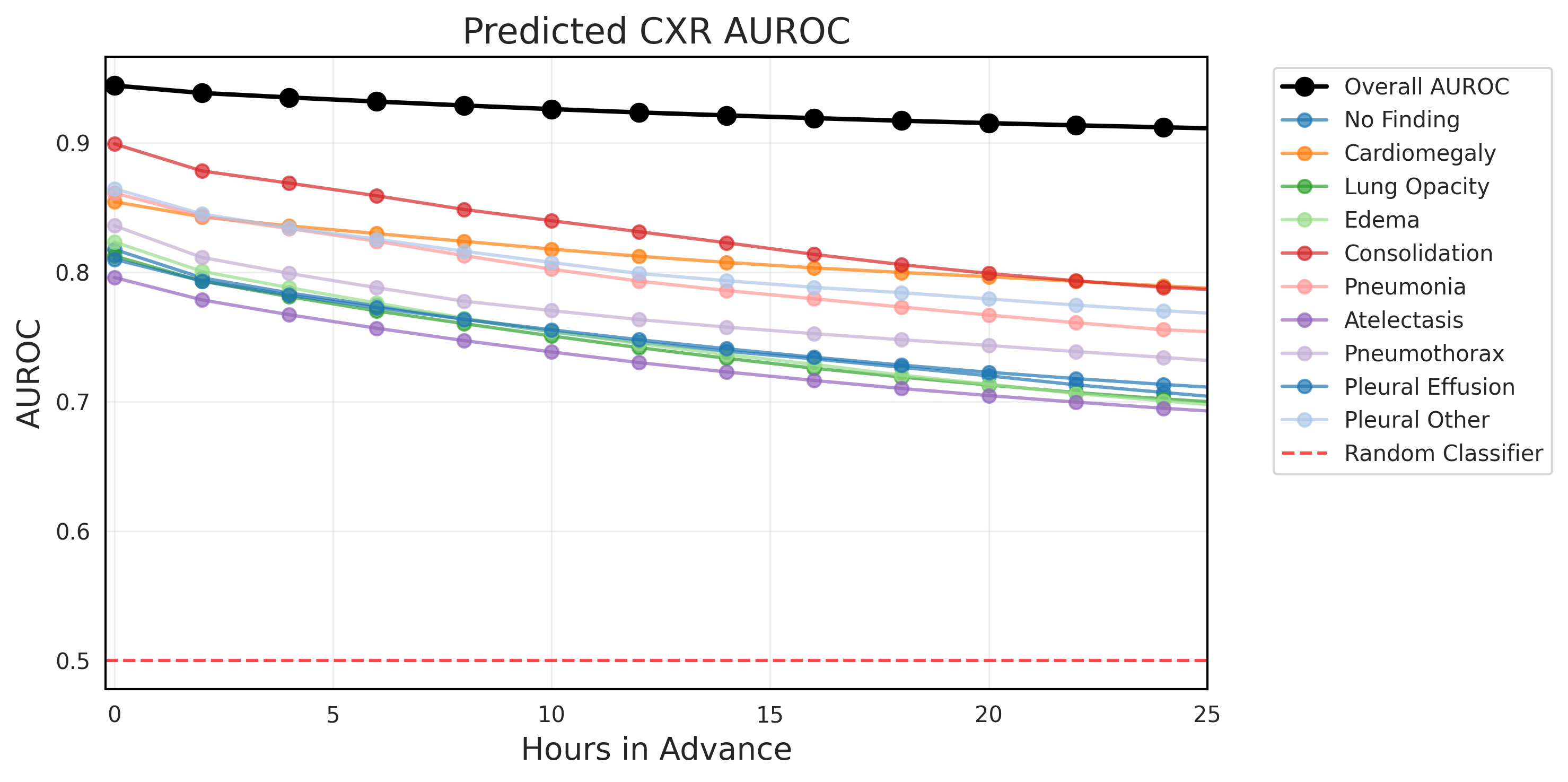}
        \caption{Baseline: Previous CXR Embeddings}
    \end{subfigure}
    \begin{subfigure}{0.48\textwidth}
        \centering
        \includegraphics[width=\textwidth]{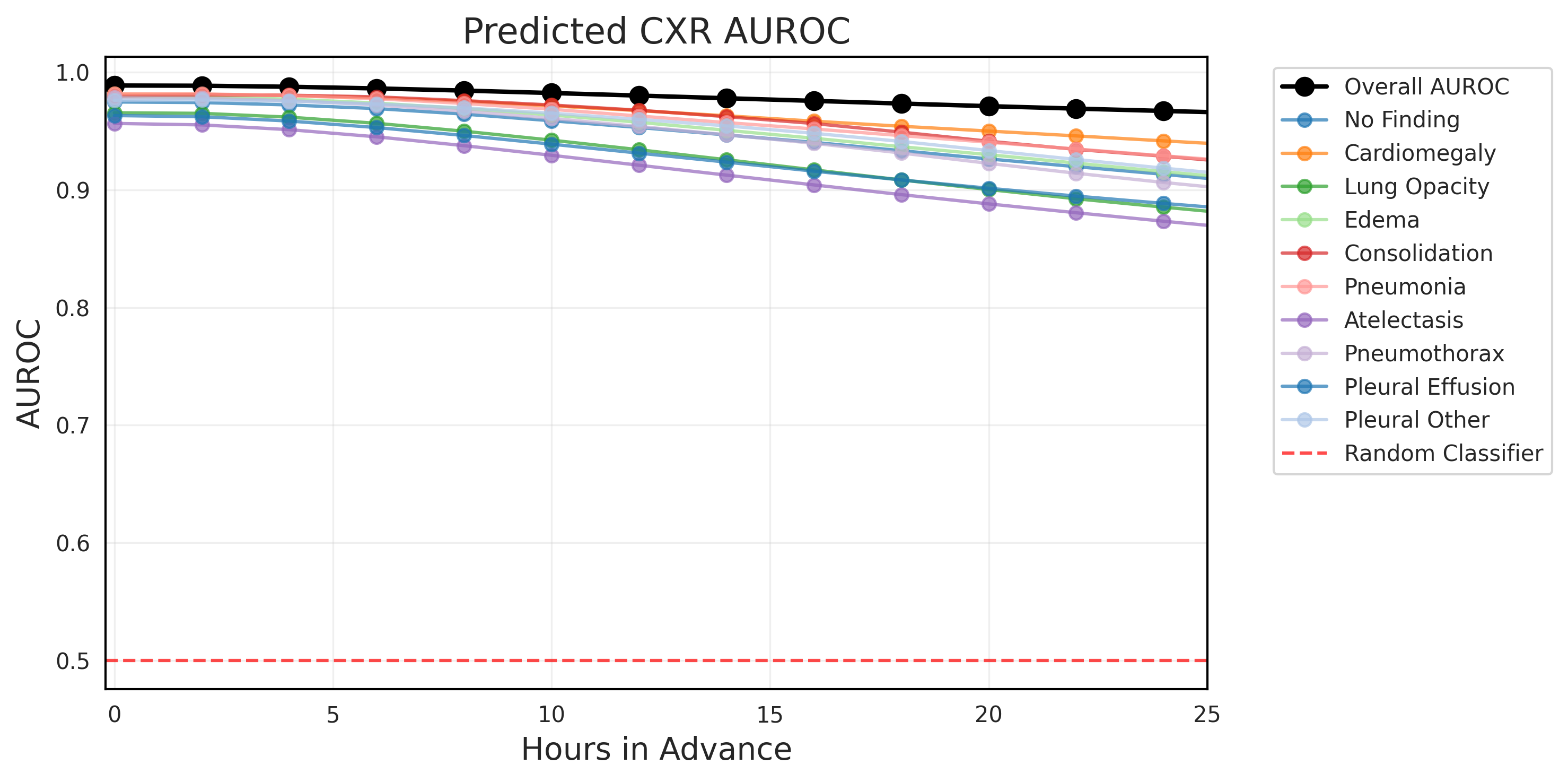}
        \caption{Predicted CXR Embeddings}
    \end{subfigure}
    
    \caption{Temporal trends in AUROC for predicting radiological findings as a function of time prior to confirmation on subsequent chest X-rays. Performance curves demonstrate the model's ability to predict findings that will be confirmed on future imaging, with prediction horizon measured in hours before documentation.}
    \label{fig:temporal-auroc}
\end{figure}

The radiographic-findings classifier (Section \ref{sec:mimic-classifier}) was used to calculate the probability of abnormal findings from CXR embeddings predicted by CXR-TFT. Following the experimental setup outlined in \cite{kyung2024towards}, the most recently recorded CXR formed the baseline for comparison. The accuracy, Area Under the Receiver Operating Curve (AUROC), and Area Under the Precision-Recall Curve (AUPRC) are reported in Table \ref{tab:results}. The 'Current Prediction' results evaluate model performance by comparing predicted labels with labels derived from interpolated target CXR trajectories. The 'Future Prediction' results assess the model by comparing predicted labels with ground truth labels from radiology reports corresponding to the subsequent CXR. 
Figure \ref{fig:roc_prc} shows the class-wise AUROC and AUPRC Curves. Figure \ref{fig:temporal-auroc}, shows the temporal variations in AUROC and AUPRC when comparing the predicted embeddings with the next recorded CXR. 

Our results show that CXR-TFT is capable of predicting radiological findings with a 95\% accuracy 12 hours before, and a 94\% accuracy 24 hours before the next CXR scan. Further, we show that predictions from CXR-TFT are a significant improvement over the baseline of the previous CXR. 

\begin{table}[h]
    \centering
    \caption{Model Performance Metrics Across Time Horizons}
    \label{tab:metrics_comparison}
    \resizebox{\textwidth}{!}{%
        \begin{tabular}{l|cc|cc|cc|cc|cc|cc|cc|cc|cc}
            \toprule
            \multirow{3}{*}{\textbf{Finding}} & \multicolumn{6}{c|}{\textbf{Current Prediction}} & \multicolumn{12}{c}{\textbf{Future Prediction}} \\
            \cmidrule{2-19}
            & \multicolumn{2}{c|}{\textbf{AUROC}} & \multicolumn{2}{c|}{\textbf{AUPRC}} & \multicolumn{2}{c|}{\textbf{Accuracy}} & \multicolumn{6}{c|}{\textbf{12-hours in advance}} & \multicolumn{6}{c}{\textbf{24-hours in advance}} \\
            \cmidrule{2-19}
            & \textbf{M} & \textbf{B} & \textbf{M} & \textbf{B} & \textbf{M} & \textbf{B} & \textbf{AUROC} & \textbf{AUROC} & \textbf{AUPRC} & \textbf{AUPRC} & \textbf{Acc} & \textbf{Acc} & \textbf{AUROC} & \textbf{AUROC} & \textbf{AUPRC} & \textbf{AUPRC} & \textbf{Acc} & \textbf{Acc} \\
            & & & & & & & \textbf{M} & \textbf{B} & \textbf{M} & \textbf{B} & \textbf{M} & \textbf{B} & \textbf{M} & \textbf{B} & \textbf{M} & \textbf{B} & \textbf{M} & \textbf{B} \\
            \midrule
            \midrule
            No Finding & 0.975 & 0.818 & 0.658 & 0.249 & 0.981 & 0.957 & 0.953 & 0.818 & 0.512 & 0.249 & 0.976 & 0.957 & 0.913 & 0.818 & 0.379 & 0.249 & 0.973 & 0.957 \\
            Cardiomegaly & 0.981 & 0.854 & 0.612 & 0.305 & 0.983 & 0.971 & 0.967 & 0.854 & 0.998 & 0.305 & 0.973 & 0.971 & 0.941 & 0.854 & 0.996 & 0.305 & 0.968 & 0.971 \\
            Lung Opacity & 0.966 & 0.812 & 0.878 & 0.673 & 0.892 & 0.832 & 0.934 & 0.812 & 0.764 & 0.673 & 0.907 & 0.832 & 0.885 & 0.812 & 0.636 & 0.673 & 0.880 & 0.832 \\
            Edema & 0.979 & 0.823 & 0.822 & 0.476 & 0.943 & 0.903 & 0.957 & 0.823 & 0.457 & 0.476 & 0.983 & 0.903 & 0.915 & 0.823 & 0.314 & 0.476 & 0.980 & 0.903 \\
            Consolidation & 0.980 & 0.899 & 0.518 & 0.285 & 0.979 & 0.970 & 0.968 & 0.899 & 0.350 & 0.285 & 0.990 & 0.970 & 0.929 & 0.899 & 0.214 & 0.285 & 0.989 & 0.970 \\
            Pneumonia & 0.981 & 0.861 & 0.445 & 0.210 & 0.981 & 0.969 & 0.963 & 0.861 & 0.499 & 0.210 & 0.982 & 0.969 & 0.929 & 0.861 & 0.373 & 0.210 & 0.979 & 0.969 \\
            Atelectasis & 0.956 & 0.796 & 0.661 & 0.380 & 0.906 & 0.827 & 0.921 & 0.796 & 0.815 & 0.380 & 0.869 & 0.827 & 0.873 & 0.796 & 0.720 & 0.380 & 0.833 & 0.827 \\
            Pneumothorax & 0.978 & 0.836 & 0.559 & 0.242 & 0.973 & 0.957 & 0.954 & 0.836 & 0.462 & 0.242 & 0.981 & 0.957 & 0.906 & 0.836 & 0.312 & 0.242 & 0.976 & 0.957 \\
            Pleural Effusion & 0.963 & 0.810 & 0.770 & 0.480 & 0.922 & 0.860 & 0.931 & 0.810 & 0.880 & 0.480 & 0.869 & 0.860 & 0.888 & 0.810 & 0.807 & 0.480 & 0.829 & 0.860 \\
            Pleural Other & 0.976 & 0.864 & 0.371 & 0.175 & 0.979 & 0.964 & 0.960 & 0.864 & 0.381 & 0.175 & 0.987 & 0.964 & 0.918 & 0.864 & 0.212 & 0.175 & 0.984 & 0.964 \\
            \midrule
            \textbf{Average} & \textbf{0.974} & \textbf{0.842} & \textbf{0.610} & \textbf{0.334} & \textbf{0.958} & \textbf{0.927} & \textbf{0.951} & \textbf{0.842} & \textbf{0.612} & \textbf{0.334} & \textbf{0.952} & \textbf{0.927} & \textbf{0.910} & \textbf{0.842} & \textbf{0.496} & \textbf{0.334} & \textbf{0.939} & \textbf{0.927} \\
            \bottomrule
        \end{tabular}%
    }
    \small \raggedright Note: M = CXR-TFT Model, B = Baseline, Acc = Accuracy.  
    \label{tab:results}
\end{table}

\section{Discussions}

With CXR-TLT, we demonstrate that modeling CXR trajectories in the latent space of a pretrained vision-language model—integrating prior CXR and clinical data—can accurately predict abnormal findings 12-24 hours before they appear on subsequent CXRs.
This methodology and the findings are novel for some critical reasons. First, there are important clinical implications of radiographical embedding prediction. By being able to estimate when a radiograph may have abnormal findings, this can accelerate clinical decision making by prompting early diagnostic imaging and subsequently clinical intervention. For example, our model may predict development of a pneumonia many hours prior to a clinical diagnosis, which may lead to earlier antibiotics and potentially decreased complications. Another strength is the simplicity of temporally aligning information from multiple irregularly sampled time-series by modeling trajectories in a continuous embedding space. The novelty of our framework is in the temporal granularity in estimating the likelihood of radiological findings on the otherwise infrequently recorded CXRs, effectively performing a super-resolution in time. 

Most previous radiological trajectory research is limited to broad categorizations (worsening, stable, improving) \cite{gourdeau2022tracking} or predicting severity outcomes (mortality, ICU readmission) \cite{duanmu2022deep} \cite{ahn2024temporal}, offering limited clinical utility as their predictions rarely influence real-time patient management and typically require current CXRs or reports. In contrast, our approach results in an actionable prediction that can be used at the bedside by clinical physicians by reflecting important physiological changes at higher temporal resolution than the relatively sparse chest x-rays alone.

The closest work to our research is the CXR generation model proposed by Kyung et al. \cite{kyung2024towards}, which uses a diffusion model to generate future chest x-rays from clinical time series data, conditioned on the most recent CXR. While groundbreaking in creating clinically actionable chest x-ray systems, our approach diverges in several key aspects. First, CXR-TLT predicts future CXRs in latent embedding space rather than pixel space, making it data and compute efficient while eliminating hallucination risks common in generative models for images- all without sacrificing clinical utility for early respiratory deterioration alerts. 
Second, our transformer model harnesses comprehensive contextual information by incorporating clinical measurements and previous CXR embeddings from admission until prediction time, enriching the model's understanding of the patient's radiological history. Third, our approach achieves hourly temporal alignment between CXR embeddings and clinical measurements, providing finer-grained monitoring capabilities.

Our work does have some limitations. First, clinical studies have demonstrated that shifting from daily CXRs to a more restrictive approach with clinically indicated imaging has resulted in similar outcomes \cite{toy2022imaging},\cite{krivopal2003utility}, \cite{clec2008daily}, \cite{graat2007elimination}.
Therefore, most ICUs today do not routinely perform daily CXRs. So, although our model can predict radiographic trajectories, the clinical significance of this cannot be determined from this study. Next, we used a single institution for our cohort, which limits the generalization of our findings. Lastly, our work focuses on a single approach to the sequence-to-sequence task but could be improved by exploring alternative model architectures like state-space models \cite{gu2023mamba}, better fusion strategies \cite{tölle2025arbitrarydataimagesfusion}, \cite{rasekh2024towards}, and more sophisticated embedding interpolation techniques.

\section{Conclusion}

 In conclusion, this work is proof of principle that a multimodal prediction model based on clinical time-series data and the latent embedding space of a pretrained vision language model can successfully predict future radiological findings. Future directions of research include further retrospective and prospective clinical studies to validate findings, exploring different model architectures, and expansion to include other data and imaging modalities. 

%
%
%
\clearpage
\bibliographystyle{splncs04}
\bibliography{mybibliography}
%

\end{document}